\pdfoutput=1
\documentclass{article}
\usepackage{arxiv}

\usepackage[utf8]{inputenc} 
\usepackage[T1]{fontenc}    
\usepackage{hyperref}       
\usepackage{url}            
\usepackage{booktabs}       
\usepackage{amsfonts}       
\usepackage{nicefrac}       
\usepackage{microtype}      
\usepackage{lipsum}		
\usepackage{graphicx}
\usepackage[numbers]{natbib}
\usepackage{doi}
\usepackage{url}
\usepackage{graphicx} 
\usepackage{adjustbox}
\usepackage{amsmath}
\usepackage{color}
\usepackage{makecell}
\usepackage{amssymb}
\usepackage{fixmath}
\usepackage{subcaption}
\usepackage{dblfloatfix}
\usepackage{multirow}

\newcommand{\x}{x}
\newcommand{\y}{y}
\newcommand{\z}{z}
\newcommand{\np}{n}
\newcommand{\points}{p}
\newcommand{\vel}{v}

\newcommand{\accel}{a}

\newcommand{\torque}{\tau}

\newcommand{\module}{m}
\newcommand{\link}{link}
\newcommand{\linkL}{r}
\newcommand{\linkD}{\Delta}
\newcommand{\linkoffset}{d}
\newcommand{\linktwist}{\alpha}
\newcommand{\jointangle}{\theta}

\newcommand{\origin}{O}
\newcommand{\controls}{\Theta}
\newcommand{\control}{C}
\newcommand{\ef}{EF}
\newcommand{\jacobian}{\textbf{J}}
\newcommand{\inertiatensor}{\textbf{I}}

\newcommand{\design}{D}
\newcommand{\config}{q}
\newcommand{\state}{S}
\newcommand{\setofobstacles}{Sph}
\newcommand{\setofwalls}{Walls}
\newcommand{\walls}{w}
\newcommand{\obstacle}{o}
\newcommand{\OC}{B}
\newcommand{\Task}{T}
\newcommand{\traj}{Traj}
\newcommand{\setpoints}{P}
\newcommand{\actuator}{act}
\newcommand{\externalload}{F}

\title{Synthesizing Modular Manipulators For Tasks With Time, Obstacle, And Torque Constraints}

\author{ Thais Campos\thanks{Corresponding author.} \\
	Sibley School of Mechanical and Aerospace Engineering\\
	Cornell University\\
	Ithaca, New York 14850\\
    Email: 
	\texttt{tcd58@cornell.edu} \\
	\And
	Hadas Kress-Gazit\\
	Sibley School of Mechanical and Aerospace Engineering\\
	Cornell University\\
	Ithaca, New York 14850\\
    Email: 
	\texttt{hadaskg@cornell.edu} \\
}

\date{}

\renewcommand{\headeright}{}
\renewcommand{\undertitle}{}

\hypersetup{
pdftitle={Synthesizing Modular Manipulators For Tasks With Time, Obstacle, And Torque Constraints},
pdfsubject={cs,robotics},
pdfauthor={Thais Campos},
pdfkeywords={Modular manipulators},
}

\begin{document}
\maketitle

\begin{abstract}
	Modular robots can be tailored to achieve specific tasks and rearranged to achieve previously infeasible ones. The challenge is choosing an appropriate design from a large search space. In this work, we describe a framework that automatically synthesizes the design and controls for a serial chain modular manipulator given a task description. The task includes points to be reached in the 3D space, time constraints, a load to be sustained at the end-effector, and obstacles to be avoided in the environment. These specifications are encoded as a constrained optimization in the robot's kinematics and dynamics and, if a solution is found, the formulation returns the specific design and controls to perform the task. Finally, we demonstrate our approach on a complex specification in which the robot navigates a constrained environment while holding an object.
\end{abstract}


\section{Introduction}
\begin{figure*}[htb]
\centering
\includegraphics[width=\textwidth]{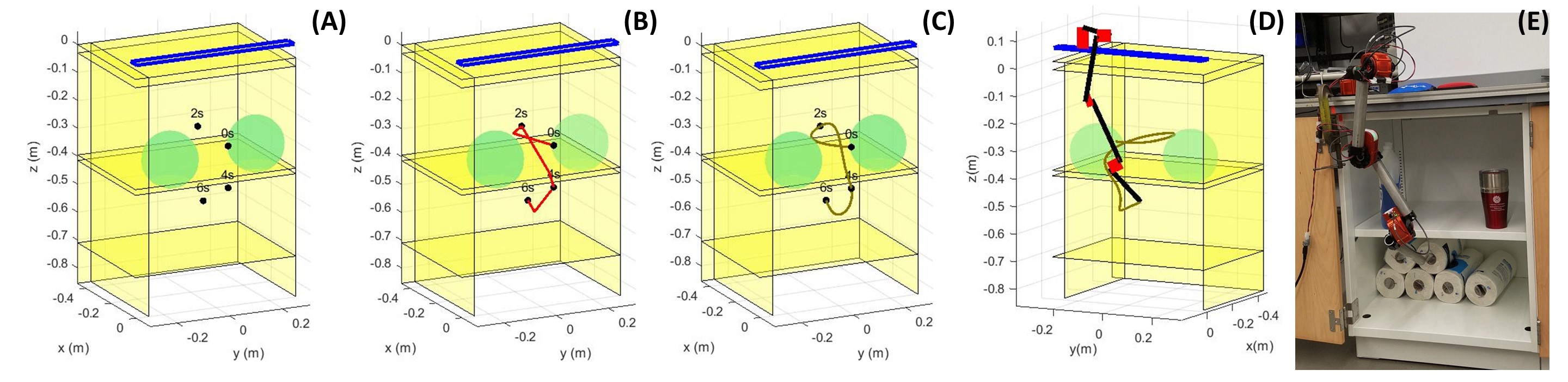}
\caption{From task definition to execution. (A) Points to be reached with their times stamps, spherical obstacles in green, planar obstacles in yellow, constraint on the origin of the manipulator in blue; (B) Connection between the task points after RRT*; (C) Trajectory over time after a polynomial fit; (D) Simulation of the resultant manipulator; (E) Hardware demonstration.}
\label{fig:fromTask2execution}
\end{figure*}

Typical manipulators can perform a set of tasks; however, they are limited by their fixed structure and actuators specifications. Modular robots, on the other hand, can be designed to achieve a specific task and adjusted to handle tasks that otherwise would be infeasible, making them more versatile and easier to repurpose. This is only possible because modular robots consist of individual components that can be reorganized or replaced to create different designs, each with its own functionality. However, manually selecting a manipulator's structure and its controls can be a difficult and error-prone process, as it involves analyzing kinematic and dynamic equations as well as deciding on discrete and continuous properties, such as number of degrees-of-freedom (DOF) and link lengths. 

In this work, we automate the design process for modular manipulators: we introduce an end-to-end system that automatically synthesizes both structure and controls of modular manipulators from high-level task specifications. Here, the task (depicted in Fig.~\ref{fig:fromTask2execution}-(A)) is a sequence of points in 3D space that the manipulator's end-effector ($EF$) must reach at specific time instants while sustaining a load applied to the $EF$, a set of  planar and spherical obstacles, and constraints on the position of the base, depicted as yellow rectangles, green spheres, and a blue rectangular region, respectively. 
Our system outputs the robot's structure, which is a serial chain manipulator with a fixed base composed of links sequentially connected by one DOF modular actuators, as seen in Fig. \ref{fig:fromTask2execution}-(E), where the red components are HEBI X-series actuators \cite{Hebi2021}. It also outputs the actuator 
control commands for the robot to be able to perform the task.

In our approach, the synthesis problem is divided into three steps: \textbf{trajectory selection}, \textbf{constrained optimization} and \textbf{verification}. In the trajectory selection step, we first use the sampling-based motion planner RRT* \cite{karaman2011sampling} to select a path, defined as a sequence of waypoints, in 3D space connecting all the points in the original task (Fig.~\ref{fig:fromTask2execution}-(B)). Then, we construct a trajectory from the selected path using polynomial interpolation (Fig.~\ref{fig:fromTask2execution}-(C)). The trajectory defines the position of the robot's end-effector over time, thus specifying its velocities and accelerations. In the second step, we solve a constrained optimization problem in the robot's kinematics and dynamics, so the manipulator can reach the points on the path individually at their respective time instants. Finally, in the verification step, we check if the previously found robot design is able to follow the trajectory while avoiding obstacles and satisfying the dynamic constraints. If our approach is able to find a solution, then it is guaranteed to be correct. However, the approach is not complete; there might exist solutions for a given task specification that our algorithm was not able to find in the allotted time. 

\textbf{Related work:} In \cite{campos2019task,van2009optimal,baykal2017asymptotically,chocron2008evolutionary,yang2000task,chen1995determining,stravopodis2021rectilinear,chung1997task,patel2015task,tabandeh2016memetic,valente2016reconfigurable,dogra2021optimal,leger1999automated,whitman2020modular,icer2017evolutionary,whitman2018task,kim1993formulation,liu2020optimizing,al2013task,singh2018modular,althoff2019effortless} explored different techniques for finding robot design such that the task---reaching a set of points in 3D---is feasible. While in \cite{campos2019task,van2009optimal,baykal2017asymptotically,chocron2008evolutionary,yang2000task,stravopodis2021rectilinear,chung1997task,chen1995determining} the authors analyzed kinematic requirements, such as dexterity \cite{van2009optimal,chocron2008evolutionary}, manipulability \cite{chung1997task} and maximization of reachable space \cite{baykal2017asymptotically}, in \cite{patel2015task,tabandeh2016memetic,althoff2019effortless,valente2016reconfigurable,leger1999automated,whitman2020modular,icer2017evolutionary,whitman2018task,liu2020optimizing,kim1993formulation,al2013task,singh2018modular,dogra2021optimal,liu2020optimizing}, the authors also incorporated torque specifications in their formulation. 

In \cite{leger1999automated}, the authors used a priority rating system in the robot's metric specifications, in which high priority items were more likely to be satisfied by the selected robot structure. The metrics included task completion, average torque applied in the actuators and completion time, ranked from most important to least. Given the prioritization formulation and the average torque calculation, the actuators torque limits might not be satisfied for the total duration of the task.

In \cite{whitman2018task,whitman2020modular,icer2017evolutionary,kim1993formulation}, the authors included static torque requirements in their approach. The total torque applied to the robot's actuator is composed of static and dynamic components. While the static torque is the one required for the manipulator to hold a position against gravity, the dynamic torque refers to the one necessary for the manipulator to perform a pre-determined motion. If for a given task, such as motion from point A to point B, the robot does not need to follow a schedule, then it can perform this motion very slowly, and the torques required in each actuator are close to the static ones. However, some tasks require the robot to reach points at certain time instants, for example when coordinating with other robots, or when moving payloads. In these cases, the robot motion must adhere to specific velocity and acceleration profiles while respecting torque limits. 
Thus, the dynamic component of the torque must be considered as well. In our approach, by specifying the $EF$ trajectory before attempting to find a manipulator structure, we include dynamics constraints in our formulation guaranteeing that, when applying the controls to the selected design, it will reach the task points at the times specified while respecting the actuators torque limits. 

Both static and dynamic components of the torque were incorporated in \cite{valente2016reconfigurable,tabandeh2016memetic,dogra2021optimal}. However, the authors only examined environments without obstacles, whose presence restricts the robot's collision-free reachable space. This restriction causes two challenges not addressed in their formulation: first, the task might require 
a more complex robot with more DOFs in order to avoid collisions; second, the path generated might be longer due to the obstacles, therefore to maintain the original timing, 
the robot might require higher torques than are supported by its actuator. 

Similar to our task definition, kinematic and dynamic specifications are also shown in \cite{althoff2019effortless,al2013task,liu2020optimizing,singh2018modular}, where the total torque is considered when finding a design for a manipulator in an environment with obstacles. However, in \cite{althoff2019effortless,al2013task,liu2020optimizing}, the authors fix the number of actuators in the manipulator design beforehand and only allow conventional values of twist angle. These restrictions create a smaller search space
that might lead to no solutions in a highly constrained environment. Moreover, this approach might select complex designs with a large number of modules when a simpler solution is sufficient and preferred. In \cite{singh2018modular} and in our approach, these restrictions are not imposed, creating structures more suitable and less complex for the task at hand. However, in \cite{singh2018modular}, the authors do not consider the robot's final trajectory in their formulation, but only the task points. When considering only certain poses for the robot in a constrained environment, the synthesis process might create structures with disconnected reachable space, as seen in \cite{campos2019task}, resulting in an infeasible task. In our approach, by simultaneously searching for a robot's design and a path that connects the task points, we guarantee that the solution is able to complete the task while satisfying all kinematics and dynamics constraints.

\textbf{The main contributions of this paper are}: (i) a framework for automated design of a manipulator structure and its controls for a task with specific timing and workspace constraints, that (ii) guarantees that the actuators do not exceed torque constraints while executing the task. We demonstrate our approach with HEBI actuators for a physical demonstration involving a complex task.

\section{Definitions}
The notation $a.b$ refers to a parameter $b$ of variable $a$. For example, $o.rad$ refers to the radius $rad$ of a spherical obstacle $o$. Continuous and discrete intervals are denoted as $[a,b]$ and $\{a,…,b\}$, respectively. Line segment, vector and vector's length whose endpoints are $u$, $v$ are denoted as $\overline{uv}$, $\overrightarrow{uv}$, and $||\overrightarrow{uv}||$, respectively. The minimum distance between two curves $h_1$ and $h_2$ or a curve $h_1$ and a point $p$ is referred to as $||h_1,h_2||$ and $||h_1,p||$, respectively. 

\subsection{Task}\label{subsec:task}


The workspace of the manipulator may include a set of spherical and planar obstacles. The set of $n_{sph}$ spherical obstacles is represented by $\setofobstacles = \{\obstacle_1,...,\obstacle_{n_{sph}}\}$, where $\obstacle$ is defined by its center $\obstacle.c \in \mathbb{R}^3$ and radius $\obstacle.rad \in \mathbb{R}^+$. The planar obstacles are rectangles defined by their vertices' coordinates and represented by $\setofwalls = \{\walls_1,...,\walls_{n_{walls}}\}$, with a total of $n_{walls}$ obstacles. Fig. \ref{fig:fromTask2execution}-(A) shows the spherical obstacles in green and the planar ones in yellow.

The set of points that the robot's $\ef$ must reach at their specific time instants while sustaining an external load is defined by $\setpoints = \{(\points_1, t_1, \externalload_1),\hdots,(\points_{\np}, t_{\np}, \externalload_{\np})\}$, where $(\points_i, t_i, \externalload_i)$ is a tuple in which $i \in \{1,...,\np\}$, $\np$ is the total number of task points, $\points_i \in \mathbb{R}^3$ is the task point whose coordinates are defined with respect to the \textit{global frame}, $t_i \in \mathbb{R}^+$ is the time stamp, and $\externalload_i \in \mathbb{R}^6$ is the vector containing the forces $[f_x,f_y,f_z]$ and moments $[M_x,M_y,M_z]$ acting on the $EF$. The first task point $\points_1$ defines the start of the task; thus, $t_1=0$. Fig. \ref{fig:fromTask2execution}-(A) shows the task points in black dots with their respective time stamps.

The set $\OC$ captures the possible positions of the origin $\origin_0$ of the manipulator. $\OC$ is parallel to the x-y plane and is specified by its diagonal coordinates, $(\OC.x_{min}, \OC.y_{min})$, $(\OC.x_{max}, \OC.y_{max})$, and height $\OC.z$. Fig. \ref{fig:fromTask2execution}-(A) shows the constraints in the origin as a rectangular blue region.

Here we define the task as  $\Task=(\setpoints,\setofobstacles,\setofwalls, \OC)$. 

\subsection{Robot Structure}\label{subsec:robot}

\begin{figure}[hbt!]
\centering
\includegraphics[scale=.35]{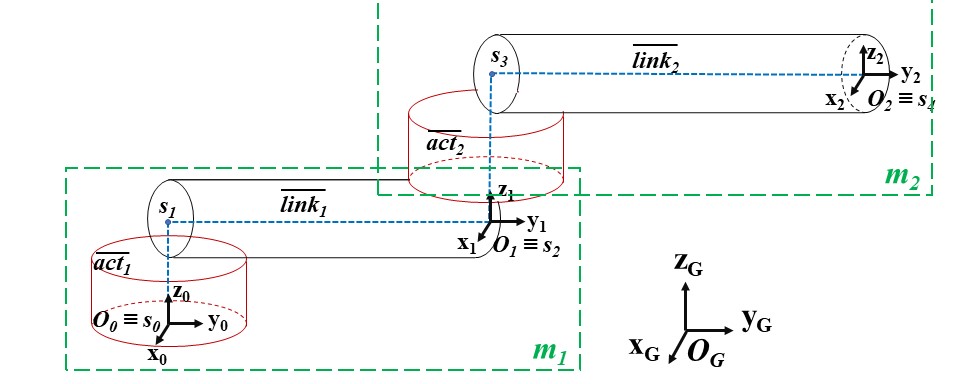}
\caption{Scheme of a 2 DOF modular manipulator. The red cylinders represent the actuators while the black ones, the links.}
\label{fig:modules}
\vspace{-3mm}
\end{figure}

We design the robot as a serial open-chain manipulator composed of modules. 
A module $\module_k$, where $k \in \{1,...,n_{DOF}\}$ and $n_{DOF}$ is the total number of modules, is composed of an actuator and a link attached to it (see Fig. \ref{fig:modules}). The actuator provides one rotational degree-of-freedom (DOF) to the structure,
while the link is a rigid tube with fixed length $\module_k.\linkL$, or $\linkL_k$ (we contract any $\module_k.var$ variable to $var_k$), and diameter $\linkD$ that connects its module to the next actuator or $\ef$. 

We attach a reference frame to each $\module_k$, with origin  $\origin_{k-1}$, $z_{k-1}$-axis in the direction of the actuator's rotation, and $x_{k-1}$ and $y_{k-1}$-axis directions defined according to the Denavit-Hartenberg (DH) convention \cite{corke2007simple}. 
Since the joints are revolute, the configuration of $\module_k$ is defined by the joint angle $\jointangle_k$. The other DH parameters are link offset $\linkoffset_k$, which is the offset along $z_{k-1}$-axis between $\origin_{k-1}$ and $\origin_{k}$, defined by the actuators dimensions, and the link twist $\linktwist_k$, which is the angle between the rotation axis of two consecutive actuators. $z_0$ is constrained to be parallel to the global z-axis $z_G$, for easier construction of the physical robot.

We define a structure $\design$ by $\origin_0$, the total number of modules $n_{DOF}$, the DH parameters $\linktwist_k \in [0,2\pi]$, and $\linkL_k \in [\linkL_0,\linkL_{max}]$ $\forall k \in \{1,...,n_{DOF}\}$, where $\linkL_0$ is the minimum length and $\linkL_{max}$, the maximum. A configuration $\config_i$ of the manipulator is the list of joint angles $\config_i=\{\jointangle_1,...,\jointangle_{n_{DOF}}\}_i$ that uniquely defines the position of all points in the system such that $EF$ reaches $p_i$. Applying the forward kinematics ($FK$) equations in $\config_i$ returns the state of the manipulator $FK(\design,\config_i)=\state_i=\{s_0,s_1,...,s_{2n_{DOF}}\}_i$. $\state_i$ is the global coordinates of the $2n_{DOF}$ points that define the endpoints of all the links in $\design$ plus the origin $\origin_0\equiv s_0$, where $s_{2n_{DOF}}\equiv\ef$ as seen in Fig. \ref{fig:modules}. Two consecutive points in $\state_i$ form a line segment that defines the spatial position of either an actuator or a link. For example, as seen in Fig. \ref{fig:modules}, the actuator component of $\module_1$ is represented by $\overline{s_0s_1}$, or $\overline{\actuator_1}$. Similarly, the first link can be represented by $\overline{s_1s_2}$, or $\overline{\link_1}$. To clarify our symbol definition,  $\state_i.\ef$ is the end-effector position given that the manipulator state is $\state_i$. 

\subsection{Controls}\label{subsec:controls}


We define the controls to be applied to $\design$ such that $\ef$ reaches $p_i$ at time $t_i$ while sustaining external load $\externalload_i, i \in \{1,...,n\}$ as $\controls_{i} = (\config_i, \dot{\config}_i, \text{T}_{i})$, where $\dot{\config_i}=\{\dot{\jointangle}_1,...,\dot{\jointangle}_{n_{DOF}}\}_i$ and $\text{T}_i = \{\torque_{1},...,\torque_{n_{DOF}}\}_i$ are the velocities and torques, respectively, applied to each actuator. 

\subsection{Problem Statement}\label{Problem Statement}
In this work, we automatically attempt to find a structure $\design$ and a sequence of controls $\control = \{\controls_{1},...,\controls_{n}\}$ such that a robotic manipulator is able to perform task $\Task$ while satisfying kinematic and dynamic constraints.

\section{Approach}
\begin{figure*}[t]
\centering
\includegraphics[width=\textwidth]{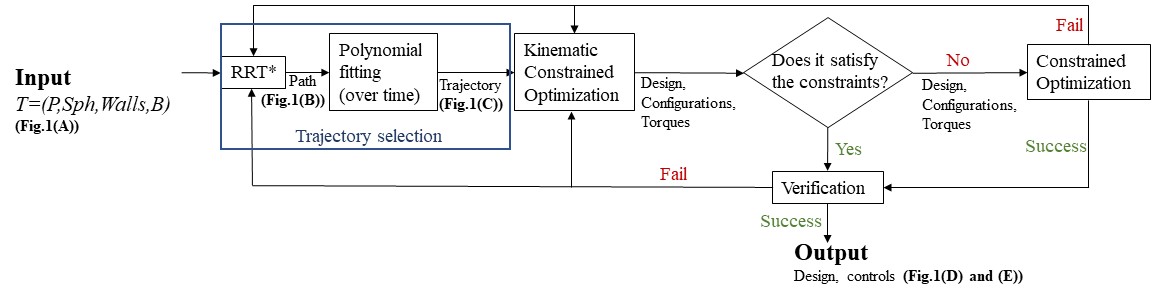}
\caption{Overview of the approach. 
}
\label{fig:approach}
\end{figure*}
\vspace{-1mm}

Fig. \ref{fig:approach} shows our overall approach. We start by using a sampling-based motion planner, RRT* (Section \ref{sec:RRT*}) to find a collision-free path for the $\ef$ in the form of waypoints. Then, through polynomial interpolation, we construct a trajectory for the robot's end-effector to follow (Section \ref{sec:polyFit}). This step is essential because we extract, from the trajectory, $\ef$ velocities and accelerations that are needed for calculating the torques applied to the actuators.

Since the existence and quality of constrained optimization solutions is highly dependent on the initial guess, we solve a simpler version of the problem with only kinematic requirements (Section \ref{subsec:kinOpt}) to generate a suitable initial guess for the full constrained optimization (Section \ref{sec:ConOpt}). We then check if the solution provided by the kinematic optimization is able to satisfy the task while meeting all kinematic (Section \ref{sec:kinCon}) and dynamic (Section \ref{sec:dyncon}) constraints. If it is, we proceed to the verification step (Section \ref{sec:verification}). Otherwise, we continue with the full constrained optimization. In the verification step, we certify that the candidate design is not only able to reach the task points but also able to follow the previously defined trajectory while complying with the torque requirements and avoiding collisions. 

Finally, if the verification is successful, the process is finished and the solution returned, otherwise, we restart the process either from the kinematic optimization using a randomly selected initial guess or from the RRT*. This restart process might be repeated until the maximum number of iterations for each step, specified by the user, is reached.

To decide on the number of actuators, we iterate over a a finite set of possible $n_{DOF}$, starting from the smallest value and adding DOFs as needed, similar to what we did in \cite{campos2019task,camposautomated2020}.


\subsection{RRT*}\label{sec:RRT*}

We start our approach with the sampling-based motion planner RRT* \cite{karaman2011sampling} to select an $\ef$ path that passes through all of the task points. RRT* is asymptotically optimal, meaning that as the number of samples grows, the path will approach the optimal Shortest2020 path. This is important for our problem since longer trajectories between two task points require higher actuator velocities and torques, which could potentially saturate the actuators.

The output of the RRT*, if successful, is a list $p^{RRT^*}$  of $\np^*$ points in the environment, $\np^* \geq \np$, that when consecutively connected by line segments, produces a collision-free path for the $\ef$ in Cartesian space, as shown in red in Fig. \ref{fig:fromTask2execution}-(B).

\subsection{Polynomial Fitting}\label{sec:polyFit}

For each new point in $p^{RRT^*}$, we assign a time instant based on its distance to the original task points, {and an external load to be sustained by the $\ef$ equals to the previous load.} For example, if a new point found by the RRT* is halfway between $p_i$ and $p_{i+1}$, then it is assigned the time instant $(t_i+t_{i+1})/2$, and the load $\externalload_i$. This process constructs a new list, $\setpoints^*$, $\setpoints \subset \setpoints^*$. 
From $\setpoints^*$, we construct a trajectory that is smooth and guarantees the actuators' velocity and acceleration continuity by interpolating piecewise polynomials over time for each coordinate, creating the trajectory $\traj=\langle \x(t),\y(t)$,$\z(t) \rangle$ (see Fig.~\ref{fig:fromTask2execution}-(C)). 
We enforce the following boundary conditions: (i) the first and second derivatives of two consecutive polynomials must be equal at the waypoint connecting them, and (ii) 
at the beginning ($t_1$) and at the end ($t_n$) of the motion, the derivatives of the polynomials are zero, so that the $\ef$ starts and ends at rest. 

When interpolating the polynomials, there are no guarantees that they do not collide with obstacles, therefore before following to the next step of the approach, we certify that $\traj$ is collision-free. We discretize it finely and check if any point on $\traj$ is inside a spherical obstacle or any two consecutive points on $\traj$ are on opposite sides of any planar obstacle. If there are collisions, we rerun the RRT* for a different path.
\vspace{-2mm}
 
\subsection{Kinematic Requirements}\label{sec:kinCon}
For the kinematic constrained optimization problem, we enforce the following constraints:

\subsubsection{Reachability} The distance between
the manipulator's $\ef$ when at state $\state_i$ and the point $p_i$ is shown in Eq. \ref{eq:reachability}. For the manipulator to successfully reach $p_i$, $h_i^R=0$.
\vspace{-1mm}
\begin{align}\label{eq:reachability}
    \forall p_i \in p^{RRT^*}, h_i^R = ||\state_i.\ef,p_i||
\end{align}

\subsubsection{Collision Avoidance} Eq. \ref{eq:collavoispher} calculates
the minimum distance between the spherical obstacles and the robot's components (links and actuators), where $\Delta$ is the components' thickness. 
\vspace{-1mm}
\begin{align}
\forall i \in \{1,...,n^*\},\forall o\in\setofobstacles,\forall k \in \{1,...,n_{DOF}\}, \nonumber\\
g_i^S=\underset{o,k}{\min}\{||\state_i.\overline{\actuator_k},o.c||-(o.rad+\Delta),\nonumber \\ ||\state_i.\overline{\link_k},o.c||-(o.rad+\Delta)\}
\label{eq:collavoispher}
\end{align}

Similarly, Eq. \ref{eq:collavoiplanar} detects if there is collision between any planar obstacles and the robot's components, where the function Intersect returns True (or 1) if two points are on opposite sides of any wall and False (or 0) if they are not.

\vspace{-5mm}
\begin{align}
\forall i \in \{1,...,n^*\},\forall\walls\in\setofwalls,\forall k \in \{1,...,n_{DOF}\}, \nonumber\\
h_i^W=\underset{\walls,k}{\max}\{\text{Intersect}(\state_i.\overline{\actuator_k},\walls), \text{Intersect}(\state_i.\overline{\link_k},\walls)\}
\label{eq:collavoiplanar}
\end{align}

Finally, Eq. \ref{eq:selfcollavoi} refers to self-collision avoidance; it calculates the minimum distance between two non-consecutive links/modules minus their thickness.

\vspace{-5.5mm}
\begin{align}
\forall i \in \{1,...,n^*\},\forall k \in \{1,...,n_{DOF}\},\nonumber \\ \forall k' \in \{k+1,...,2n_{DOF}-1\},\nonumber \\
g_i^{SC}=\underset{k,k'}{\min}(||\state_i.\overline{s_{k-1}s_{k}},\state_i.\overline{s_{k'}s_{k'+1}}||-2\Delta)
\label{eq:selfcollavoi}
\end{align}

For the robot to successfully avoid any collisions when reaching $p_i$, we require $g_i^S\geq0, h_i^W=0,$ and $g_i^{SC}\geq0$.

\subsubsection{Origin} The manipulator's origin position is bounded by $\OC$, as shown in Eq. \ref{eq:origin}. 
\begin{align}\label{eq:origin}
    \OC.x_{min} \leq \origin_0.x \leq \OC.x_{max}, \OC.y_{min} \leq \origin_0.y \leq \OC.y_{max}, \nonumber \\ \origin_0.z\equiv \OC.z
\end{align}

\subsubsection{Link Length} Eq. \ref{eq:linkLength} enforces that the previous link length is always equal to or smaller than the next. This constraint might contradict the usual manipulator design, in which longer links are positioned closer to the base together with heavier and more powerful actuators. However, here we are considering identical modules. Smaller link lengths closer to the base reduces the distance from the actuators closer to the $\ef$ to the actuators closer to the base. Since the actuators are the heaviest components (except, possibly, for the load), reducing the link lengths close to the base reduces the torque required there. 
Moreover, shorter arms contribute for better accuracy and stability and, thus, improving the robot's behavior. 
If the modules are not identical, this constraint can be removed.

\vspace{-5mm}
\begin{align}\label{eq:linkLength}
    \forall k \in \{1,...,n_{DOF}\},  \linkL_{k-1} \leq \linkL_k \leq \linkL_{max}
\end{align}

\subsection{Dynamic Requirements}\label{sec:dyncon}



We use the Recursive Newton-Euler Algorithm (RNEA) \cite{siciliano2016springer} to calculate the joint torques required for a serial manipulator to perform a trajectory. Next, we present how we compute the necessary parameters to perform the RNEA.

\subsubsection{EF Velocity and Acceleration}

We calculate the velocity $\boldsymbol{v_i}$ and acceleration $\boldsymbol{\accel_i}$ at the $\ef$, with respect to the base frame (located at $\origin_0$), required when reaching $p_i \in \setpoints^*$ by taking the first and second derivative of $\traj$ and evaluating them at time stamp $t_i$, $\boldsymbol{v_i}=[\dot{x}(t_i); \dot{y}(t_i);\dot{z}(t_i)]$ and $\boldsymbol{\accel_i}=[\ddot{x}(t_i); \ddot{y}(t_i);\ddot{z}(t_i)]$. Since we do not control the orientation of the $\ef$, we only consider the linear components of the robot's spatial velocity.  Therefore, we eliminated the angular velocity (usually, referred to as $\omega_x, \omega_y, \omega_z$) and removed the corresponding rows in the jacobian matrix. 

\subsubsection{Actuators' Velocity and Acceleration}

The relation between $\boldsymbol{\vel_i}$, $\boldsymbol{\accel_i}$, $\dot{\config_i}$, and $\ddot{\config_i}$ is described by the manipulator's geometric Jacobian $\jacobian_i \in \mathbb{R}^{3\times n_{DOF}}$ \cite{siciliano2010robotics}, and shown in Eqs. \ref{eq:vJq} and \ref{eq:jointaccel}, where $\jacobian_i^{-1}$ is the pseudo-inverse matrix of $\jacobian_i$.

\vspace{-3mm}
\begin{equation}\label{eq:vJq}
    \dot{\config_i}=\jacobian_i^{-1}\boldsymbol{\vel_i}
\end{equation}

Taking the time derivative of Eq. \ref{eq:vJq} yields Eq. \ref{eq:jointaccel}.
\vspace{-1mm}
\begin{equation}\label{eq:jointaccel}
    \ddot{\config_i} = \jacobian_i^{-1}(\boldsymbol{\accel_i} - \dot{\jacobian_i}\dot{\config_i})
\end{equation}
\vspace{-5mm}

\subsubsection{Inertia Tensor} The module 
$\module_k$, $k \in \{1,...,n_{DOF}\}$, is a compound object composed of an actuator and a link. To calculate its inertia tensor $\inertiatensor_k\in \mathbb{R}^{3\times 3}$, we model the actuator as a solid uniform cylinder whose mass, height, and radius are approximated by the measurements provided by the manufacturer. The link is a hollow cylinder with length $r_k$, density $\rho$, and external and internal radius $rad_2$ and $rad_1$, respectively. $\inertiatensor_k$ is, then, the sum of the actuator and the link inertia tensors calculated with respect to the frame attached to $\module_k$, whose position and axis are defined following DH convention (Section \ref{subsec:robot}). 

\subsubsection{RNEA}

The RNEA has two steps: a \textit{forward} and a \textit{backward} recursion. In the forward recursion, we progressively calculate the modules' velocities and accelerations with respect to the base frame, using the previously calculated parameters (actuators poses, velocities, and accelerations), as well as the velocity and acceleration of the base frame, which are referred to as initial conditions. In our formulation, we consider that the base frame is static.

Then, in the backward step, we use the results from the forward recursion, the modules' inertia tensors, and the force and moment exerted at $\ef$ ($\externalload_i$) to recursively compute the forces and torques required for each actuator. For a detailed explanation of the algorithm, refer to \cite{siciliano2010robotics}. Finally, with the torques required for all actuators at time stamps $t_i, \forall i \in \{1,...,\np^*\}$, we define Eq. \ref{eq:dynamic_constraints}, which calculates the minimum difference between the maximum torque $\torque_k^{max}$ allowed and the torque required in the actuators at specific time instants $\torque_{k,i}$. If $g_i^{\torque}\geq0$, then the actuators satisfy the torque limits. In Eq. \ref{eq:dynamic_constraints}, we add $\beta > 1$ as a safety factor used to protect the actuators from saturation. From our experience with the physical modules, nonlinearities such as overshoot
and sample time may cause slight deviations from the desired trajectories. Therefore we add $\beta$ to create structures and controls that are more constrained and thus more robust to such unmodeled physical phenomena.  

\vspace{-5mm}
\begin{align}\label{eq:dynamic_constraints}
\forall i \in  \{1,...,\np^*\}, \forall k \in \{1,...,n_{DOF}\},  \nonumber \\
g_i^{\torque} = \underset{k}{\min} (\torque_k^{max}-|\beta\torque_{k,i}|)
\end{align}

\subsection{Kinematics Constrained Optimization}\label{subsec:kinOpt}

The kinematics constrained optimization is defined in Eq. \ref{eq:kinopt}. The cost function $f^{kin}$ includes collision avoidance objectives (Eqs. \ref{eq:collavoispher},\ref{eq:collavoiplanar},\ref{eq:selfcollavoi})  and the constraints include reaching the task points, origin placement, and link lengths constraints (Eqs. \ref{eq:reachability},\ref{eq:origin},\ref{eq:linkLength}). Since we view the kinematic optimization as providing an initial solution, we encode the collision avoidance as soft constraints by placing them in the cost function. This makes the optimization problem easier to solve. This might result in solutions that are not collision free; however, the design returned might still be able to reach the task points using different configurations that are collision free. We decided to maintain the reachability requirement as a hard constraint so that the structure is guaranteed to reach the task points. 
Given a solution, we first check if any configuration returned collides with an obstacle. If it does, we then search for a new configuration that is collision-free using the inverse kinematics (IK) equations. If one is found, we replace it in the solution.  

While one can choose to add weights to the different components in the cost function, we set all weights equal to $\pm1$ because they are all in comparable ranges and have the same importance (a solution with self collisions is as inadequate as one that collides with an obstacle). 

\vspace{-4mm}
\begin{equation}\label{eq:kinopt}
\begin{gathered}
\text{min }f^{kin} = \sum_{i=1}^{\np^*} (-g_i^S+h_i^W-g_i^{SC}),\\ 
\text{Subject to: } h_i^R=0, \text{and Eqs. }\ref{eq:origin}, \ref{eq:linkLength}
\end{gathered}
\end{equation}

For some tasks, the kinematic optimization is enough to yield a solution that also satisfies the dynamic requirements. For example, for tasks in which there are no loads applied in the $\ef$ and the robot design is compact (small link lengths and small number of DOFs), the torques required for the actuators may be small. Therefore, we first check whether the solution provided by this optimization satisfies both kinematic and dynamic requirements by calculating if $\forall i \in \{1,...,n^*\}, h_i^R=0, h_i^W=0, g_i^S \geq 0,g_i^{SC}\geq0$, and $g_i^{\torque}\geq0$. If it does, we continue to the verification step, otherwise, this solution is used as initial guess for the full  constrained optimization. 

\subsection{Constrained Optimization}\label{sec:ConOpt}

The full constrained optimization guarantees that the design selected reaches all task points while avoiding collisions and satisfying torque requirements, as shown in Eq. \ref{eq:opt}. We use a different cost function than in Eq. \ref{eq:kinopt} in order to modify solutions returned by the kinematic optimization which might not satisfy the dynamic requirements.

\begin{equation}\label{eq:opt}
\begin{gathered}
\text{min }f' = \sum_{i=1}^{\np^*} -g_i^{\torque},\\ 
\text{Subject to: } h_i^R=0, h_i^W=0, \\ g_i^S \geq 0,g_i^{SC}\geq0, g_i^{\torque}\geq0,
\text{and Eqs. }\ref{eq:origin}, \ref{eq:linkLength}
\end{gathered}
\end{equation}

\subsection{Verification}\label{sec:verification}

The output of the optimization is a design $\design$ and its controls $\controls_{i}, \forall i \in \{1,...,\np^*\}$, required to reach the points on the path at the time instants in $\setpoints^*$. In the verification step, we check if $\design$ can not only reach these points but can also follow $\traj$ while satisfying all the constraints. For that, we first further discretize $\traj$ into $n^{\dagger}>n^*$ points $p^{\dagger}$. Then, we solve the inverse kinematics (IK) for each point $p_{i^{\dagger}},\forall i{^{\dagger}} \in \{1,...,n^{\dagger}\}$, to find the joint angles such that $\ef$ reaches the desired point while avoiding collisions. If the IK provides a solution, we check if the dynamic constraints are satisfied and if the continuity of the motion is maintained by applying Eq. \ref{eq:continuity}, where $\config_{i^{\dagger}}$ and $\config_{i^{\dagger}+1}$ are consecutive configurations and $\epsilon$ is a small threshold. The continuity constraint is necessary because there might exist several solutions for the IK problem and we are looking for solutions in which $\config_{i^{\dagger}+1}$ is similar to $\config_{i^{\dagger}}$ to ensure that the robot's $\ef$ does not deviate from the trajectory. The value we use for $\epsilon$ is max$(\dot{\config}_{i^{\dagger}},\dot{\config}_{i^{\dagger}+1})dt$, where $dt$ is the time interval and $\dot{\config}_{i^{\dagger}}$ and $\dot{\config}_{i^{\dagger}+1}$ are calculated from Eq. \ref{eq:vJq}, since it constraints the maximum possible distance (in radians) traveled by the actuator during that interval.

\begin{equation}\label{eq:continuity}
\begin{gathered}
\forall i^{\dagger}\in\{1,...,\np^{\dagger}\},
g_{i^{\dagger}}^{\torque}\geq0 \wedge |\config_{i^{\dagger}}-\config_{i^{\dagger}+1}|\leq \epsilon
\end{gathered}
\end{equation}

\section{Results}
\begin{figure}[h]
\centering
\includegraphics[scale=.4]{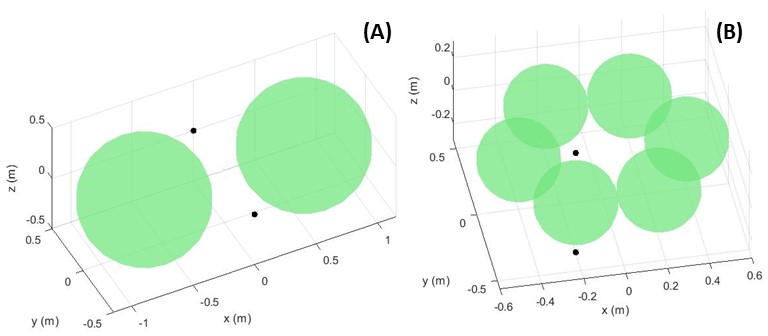}
\caption{Tasks descriptions used for comparison. The green spheres are the obstacles and the black dots the task points. The origin of the manipulator 
 is constrained to be at the origin of the global frame.
 }
\label{fig:examples}
\end{figure}

\begin{figure*}[bth!]
\centering
\includegraphics[scale=.31]{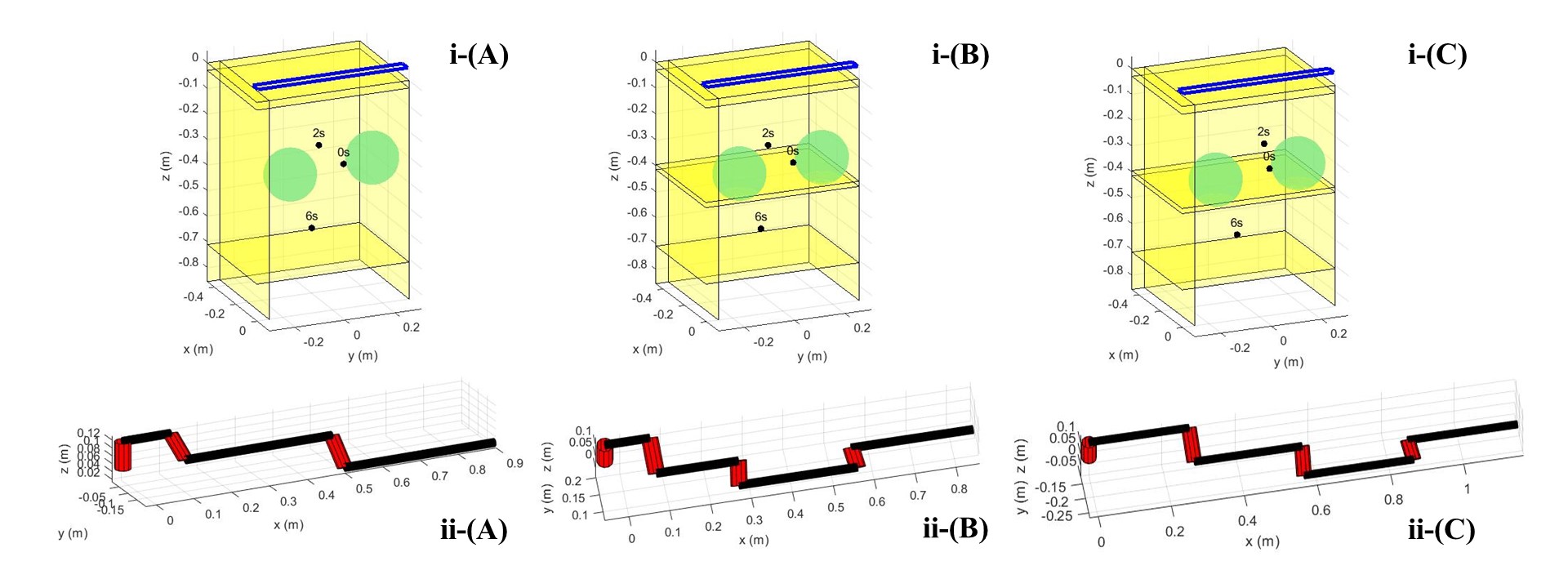}
\caption{Three different task definitions (i-A to i-C), and the corresponding returned manipulator design (ii-A to ii-C).}
\label{fig:diffdesign}
\end{figure*}

We implemented our approach in MATLAB R2020b and solved the constrained optimization problem using Sequential Quadratic Programming (SQP) with the \textit{fmincon} function. We used the RNEA algorithm from The Robotics Toolbox for MATLAB (RTB) \cite{corke2017robotics}. To calculate the distances between line segments and the collision check between line segment and planar obstacles, we used the algorithms in \cite{Shortest2020} and \cite{geom3d2020}, respectively. We implemented the RRT* algorithm based on the one in \cite{rrtmatlab2019}.

For all the simulations presented in this section, our approach took a few hours (from around 2 for tasks in Fig. \ref{fig:diffdesign} to 5 for task in Fig. \ref{fig:constrained}) to return a solution. Since we do not fix the number of DOFs beforehand, our approach might take a longer time to return a solution compared to approaches in which the number of components is fixed, as in \cite{althoff2019effortless,al2013task,liu2000performance}. However, we might return less complex and more compact structures that are easier to construct and control.

\subsection{Comparison With Other Approaches}

\begin{table}[bt!]
\caption{Comparison of the number of solutions obtained by our approach and by the kinematic-only approach for the tasks shown in Fig. \ref{fig:examples}.}
\centering
\setlength\tabcolsep{3pt}
\begin{tabular}{ |c|c|c|c|c| } 
 \hline
     & \multicolumn{2}{c|}{Example A \# Success (Fig. \ref{fig:examples}-A)} & \multicolumn{2}{c|}{Example B \# Success (Fig. \ref{fig:examples}-B)}\\
 \hline
   DOF  &  \thead{Kinematics req. opt. -\\ Trajectory search}  & \thead{Our\\ approach} & \thead{Kinematics req. opt. -\\ Trajectory search} & \thead{Our\\ approach}\\
 \hline
 2 & 22 - 0 & 0 & 0 - - & 0 \\
 \hline
 3 & 43 - 33 & 8 & 39 - 26 & 8 \\
 \hline
 4 & 61 - 47 & 65 & 44 - 35 & 42 \\
 \hline
\end{tabular}
\label{tab:resultsComparison}
\end{table}

To enable the joint synthesis of design and control from tasks with dynamic constraints, we must create a trajectory over time for the robot's end-effector to follow \textit{before} choosing a robot design.  Moreover, the candidate design needs to be able to follow the entire trajectory (checked in the Verification step), for it to be a valid solution. On the other hand, approaches that only consider kinematic or static dynamic requirements might find a trajectory for the robot to follow \textit{after} deciding on a design, by applying an off-the-shelf motion planner. 

In this section, we compare our formulation to our prior work with only kinematic requirements~\cite{campos2019task}. Here, we are interested in comparing the total number of solutions provided by each approach for the same task specification, and the running time. In order to make this analysis more adequate, we eliminate the torque constraints from our formulation. Thus, we are only evaluating how constructing the $\ef$ trajectory before (this paper) or after (prior work) choosing a robot design affects the results.

For the comparison, we implemented a kinematic-only approach adapted from our own kinematic constrained optimization in Section \ref{sec:kinCon} and inspired by \cite{campos2019task}. After a solution is found by the optimization, we run the RRT* in joint space to find a collision-free trajectory for the manipulator. The task descriptions used for comparison are shown in Fig.~\ref{fig:examples} and extracted from \cite{campos2019task}, where the kinematic-only approach synthesized a manipulator's structure with a disconnected workspace, making it impossible for the RRT* to find a trajectory for the robot to follow. 

The strategy used in \cite{campos2019task} to produce a robot design in such cases was to ask for human input.  The approach required the user to add a new task point, forcing the constrained optimization to select a design whose workspace was fully connected. In our formulation, by including the sequence of waypoints returned by the RRT* into our optimization, we can select robots that are more likely to have a fully connected workspace, not requiring user involvement. Here, we do not request user input for any of the approaches.

We ran both approaches 100 times for each  task description shown in Fig.~\ref{fig:examples} and for each number of actuators in $n_{DOF} =\{2,3,4\}$. 
The initial guesses were randomly selected but identical for both approaches. 

The number of solutions provided by each formulation is shown in Table~\ref{tab:resultsComparison}. For both tasks A and B, for a $n_{DOF}=2$, the kinematic-only optimization is able to synthesize a design (22 of 100) for task A. However, the RRT* was unable to find a trajectory in the allotted time which indicates that the workspace may be disconnected.
This corroborates the findings in \cite{campos2019task}. Our approach is also unable to find a solution for $n_{DOF}=2$, which is expected since a 2 DOF robot is underactuated 
and, thus, unlikely to be able to follow an arbitrary trajectory like the ones returned by RRT*. As $n_{DOF}$ increases, our approach is able to find more solutions than the kinematic-only approach because our optimization returns more designs with a connected workspace that are able to follow the required trajectories.

Our approach takes on average 2.6x longer to provide a solution (19.05s compared to 7.30s) for the tasks studied. This is expected, since we include all $\np^*\geq\np$ waypoints in $p^{RRT^*}$ in the optimization, thus creating more constraints. 

\subsection{Tasks and Designs}

\begin{figure}[htb]
\centering
\includegraphics[scale=.5]{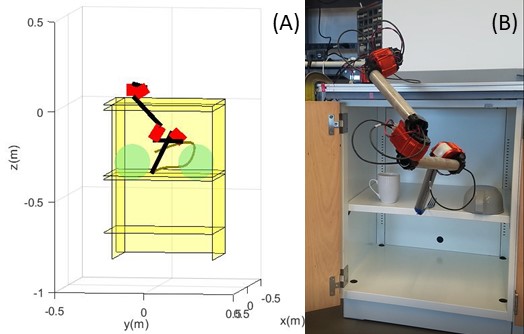}
\caption{Simulation (A) and physical robot (B) synthesized from the task shown in Fig. \ref{fig:diffdesign}-i-(C). 
}
\label{fig:simulation_real}
\end{figure}

In this section we show how a similar task with slight modifications results in different designs, showcasing the advantage of using such design automation techniques for a variety of tasks. 

We examined three slightly different task descriptions, shown in Fig. \ref{fig:diffdesign}. The manipulator's goal is to reach a point above the middle shelf inside the cabinet at $t_2=2s$, grab a payload of 33g and carry it to the bottom shelf without colliding with any wall, shelf or spherical obstacle. The task ends at $t_3=6s$. For $n_{DOF}=2$, we require $r_{max}=0.6$m, for $n_{DOF}=3$, $r_{max}=0.4$m, and for $n_{DOF}=4$, $r_{max}=0.3$m, thus maintaining a maximum span of 1.2m, regardless of the number of actuators. 
In the task presented in Fig. \ref{fig:diffdesign} i-(A), the cabinet does not have the middle shelf, which was added in task i-(B). In task i-(C), the task point to be reached at $t_2=2s$ is moved to be behind one of the spherical obstacles.  Even with a similar set of tasks, our approach was able to synthesize a diverse set of designs as seen in Fig. \ref{fig:diffdesign}-ii-(A) to (C). For task i-(A), the environment is less constrained without the middle shelf, allowing a robot with $n_{DOF}=3$ to fulfill the task; however for tasks containing the added obstacle, i-(B) and i-(C), an additional DOF is needed. Moreover, for task i-(C), the second point to be reached is more difficult to reach when compared to task i-(B). Thus, the resultant manipulator needs a longer span, as seen in Fig. \ref{fig:diffdesign}-ii-(B) and ii-(C).

\subsection{Simulation and Demonstration}

\begin{figure*}[htb]
\centering
\includegraphics[scale=.5]{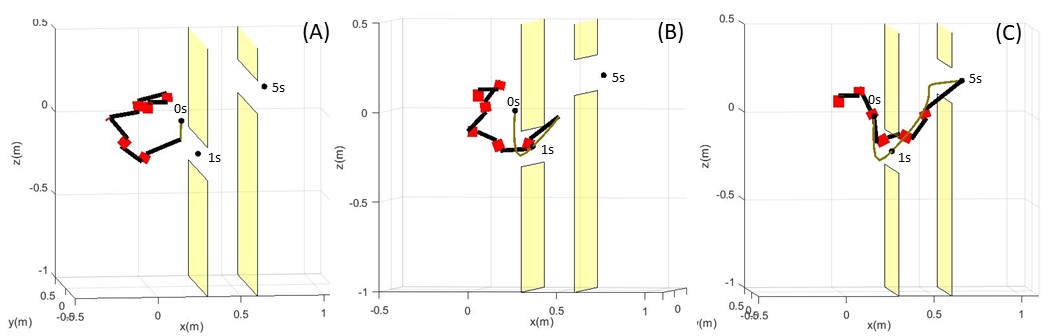}
\caption{Simulation results for a task in a highly constrained environment (A) at the beginning, (B) middle, and (C) end of the task. The original task points (black dots) and the timings are also shown. 
}
\label{fig:constrained}
\end{figure*}

To show that our approach is able to handle highly constrained environments and produce solutions with varied DOFs, we specified the task shown in Fig. \ref{fig:constrained}, in which the manipulator must travel through narrow passages, while reaching certain points at specific times (see Fig. \ref{fig:constrained}(A)). Given the constrained environment, the resultant manipulator is redundant, having 6 DOFs, which gives it the required dexterity to navigate between the walls.  

\begin{figure}[h]
\centering
\includegraphics[scale=.55]{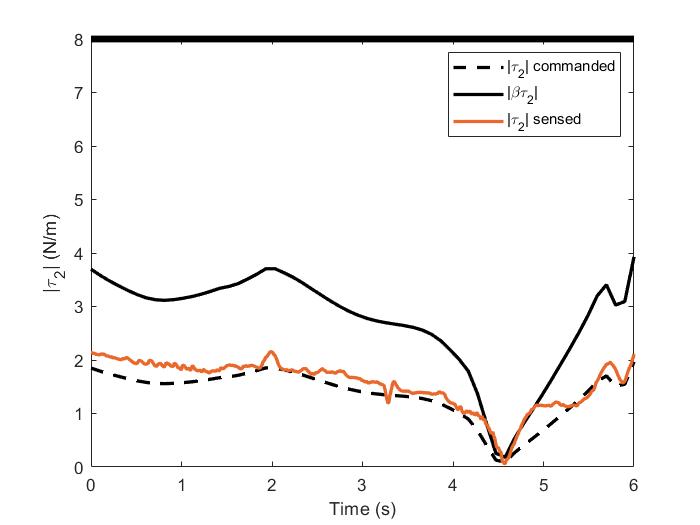}
\caption{Absolute value of torque $\tau_2$ calculated and commanded to module $k=2$ (dashed black curve), as well as, $|\beta\tau_2|$ with $\beta=2$ (black curve), and absolute value of torque sensed in the actuator during the duration of the task (orange). For the satisfaction of the torque constraint, the curve $|\beta\tau_2|$ needs to be below the $\tau^{max}$, which is assumed to be 8 N/m (black thick line).
}
\label{fig:torqueact2}
\end{figure}

In order to demonstrate the feasibility of the physical implementation of our approach, we implemented the solution found for the task in Fig. \ref{fig:diffdesign}-i-(C). The constructed robot is shown in Fig. \ref{fig:simulation_real}-(B)). The links are aluminum tubes with 31.75 mm diameter and length according to the synthesis results. Mounting brackets manufactured by HEBI Robotics (black devices showed in Fig. \ref{fig:simulation_real}-(B)) fix the actuators to the links. Fig. \ref{fig:simulation_real}-(A) also shows the simulated robot. We used an open-loop controller commanding simultaneous position, velocity and torque control for the physical robot and a magnet as the end-effector to attract and secure the payload (a fork) at $t_2=2s$. Fig. \ref{fig:torqueact2} shows the absolute value of the torque $\tau_2$ calculated and commanded to the module $k=2$, as well as, $|\beta\tau_2|$ with $\beta=2$, and the torque sensed at the actuator during the whole trajectory. We chose to plot the torques of this actuator because it endures a higher applied torque than the others as it carries more load against gravity. We considered  $\tau^{max}=8$N/m. As seen in Fig. \ref{fig:torqueact2}, the torque sensed in the actuator (orange curve) is similar to the torque commanded (dashed black curve), even though it shows some jittering possibly due to the nonlinearities inherent in physical systems, such as overshoot and hysteresis. Due to these discrepancies, we can see that around $t=4.5$s, when the torque assume small values,  the sensed torque surpass $|\beta\tau_2|$ by a negligible amount
showing the need of the safety factor for more robust structure and controls. A video accompanying this paper shows the manipulator performing the task.

\section{Conclusions and Future Work}

\textbf{Summary.} In this work, we presented a framework for automated synthesis of a serial chain modular manipulators' design and controls based on a task, introducing an end-to-end method from task description to physical implementation. Our task formulation includes a set of points in 3D space that must be reached at predefined time instants while the robot's end-effector sustains a specified load. We demonstrated that our approach synthesizes structures tailored to the task at hand and are, therefore, less complex and more compact than commercial prefabricated manipulators that contain a fixed number of DOFs. Finally, we show the feasibility of our solutions by physically implementing a manipulator whose task involved securing an object and traveling inside a cabinet, a highly constrained environment, while complying with the actuators' torque limits.

\textbf{Practical considerations:} While the physical robot successfully achieved the desired task, there might be differences between the synthesized trajectory and the performed one. This is due to several factors. First, small errors in the robot construction and placement modifies its kinematics and dynamics and, therefore, its behavior. Second, loss of accuracy characteristic of open-loop controllers (like the one used here) introduces small errors in the system that builds up with time. Checking the robot's design parameters after construction and introducing a feedback loop, with integral and derivative terms into the controller might reduce these discrepancies and improve the robots' behavior.

\textbf{Future Work.} There are a few directions we would like to explore in future work. First, if our approach provides a solution, the physical robot might only accomplish the task if its manufacturing errors, and its actuators' accuracy are within an acceptable range. Thus, we will explore techniques for the robot's self-verification, as in \cite{althoff2019effortless}, so it can check if its design parameters are similar enough to the ones from the nominal results. If they are different, it will recalculate the controls, if possible, so the task is still feasible. Second, our formulation assumes the robot structure to be a serial chain with fixed origin, which constrains the family of functions it can perform. We will extend our work to include more complex structures, such as parallel manipulators and robots that combine both locomotion and manipulation. Finally, we will include feedback to the user for infeasible tasks, as in \cite{campos2019task}. For example, this feedback could be suggestions to eliminate a task point or to modify some specifications, such as to increase the time instants to allow for smaller torques applied to the actuators.
 
\section*{Acknowledgments}
This work was funded by NSF CNS-1837506.

\bibliographystyle{unstr}

\begin{thebibliography}{10}

\bibitem{Hebi2021}
Hebi robotics.
\newblock Jun. 17, 2021. [Online].

\bibitem{karaman2011sampling}
Sertac Karaman and Emilio Frazzoli.
\newblock Sampling-based algorithms for optimal motion planning.
\newblock {\em The international journal of robotics research}, 30(7):846--894,
  2011.

\bibitem{campos2019task}
Thais Campos, Jeevana~Priya Inala, Armando Solar-Lezama, and Hadas Kress-Gazit.
\newblock Task-based design of ad-hoc modular manipulators.
\newblock In {\em 2019 International Conference on Robotics and Automation
  (ICRA)}, pages 6058--6064. IEEE, 2019.

\bibitem{van2009optimal}
EJ~Van~Henten, DA~Van’t~Slot, CWJ Hol, and LG~Van~Willigenburg.
\newblock Optimal manipulator design for a cucumber harvesting robot.
\newblock {\em Computers and electronics in agriculture}, 65(2):247--257, 2009.

\bibitem{baykal2017asymptotically}
Cenk Baykal and Ron Alterovitz.
\newblock Asymptotically optimal design of piecewise cylindrical robots using
  motion planning.
\newblock In {\em Robotics: Science and Systems}, 2017.

\bibitem{chocron2008evolutionary}
Olivier Chocron.
\newblock Evolutionary design of modular robotic arms.
\newblock {\em Robotica}, 26(3):323--330, 2008.

\bibitem{yang2000task}
Guilin Yang and I-Ming Chen.
\newblock Task-based optimization of modular robot configurations: minimized
  degree-of-freedom approach.
\newblock {\em Mechanism and machine theory}, 35(4):517--540, 2000.

\bibitem{chen1995determining}
I-Ming Chen and Joel~W Burdick.
\newblock Determining task optimal modular robot assembly configurations.
\newblock In {\em proceedings of 1995 IEEE International Conference on Robotics
  and Automation}, volume~1, pages 132--137. IEEE, 1995.

\bibitem{stravopodis2021rectilinear}
NA~Stravopodis and VC~Moulianitis.
\newblock Rectilinear tasks optimization of a modular serial metamorphic
  manipulator.
\newblock {\em Journal of Mechanisms and Robotics}, 13(1), 2021.

\bibitem{chung1997task}
Wan~Kyun Chung, Jeongheon Han, Youngil Youm, and SH~Kim.
\newblock Task based design of modular robot manipulator using efficient
  genetic algorithm.
\newblock In {\em Proceedings of International Conference on Robotics and
  Automation}, volume~1, pages 507--512. IEEE, 1997.

\bibitem{patel2015task}
Sarosh Patel and Tarek Sobh.
\newblock Task based synthesis of serial manipulators.
\newblock {\em Journal of advanced research}, 6(3):479--492, 2015.

\bibitem{tabandeh2016memetic}
Saleh Tabandeh, William Melek, Mohammad Biglarbegian, Seong-hoon~Peter Won, and
  Chris Clark.
\newblock A memetic algorithm approach for solving the task-based configuration
  optimization problem in serial modular and reconfigurable robots.
\newblock {\em Robotica}, 34(9):1979--2008, 2016.

\bibitem{valente2016reconfigurable}
Anna Valente.
\newblock Reconfigurable industrial robots: A stochastic programming approach
  for designing and assembling robotic arms.
\newblock {\em Robotics and Computer-Integrated Manufacturing}, 41:115--126,
  2016.

\bibitem{dogra2021optimal}
Anubhav Dogra, Srikant Sekhar~Padhee, and Ekta Singla.
\newblock An optimal architectural design for unconventional modular
  reconfigurable manipulation system.
\newblock {\em Journal of Mechanical Design}, 143(6), 2021.

\bibitem{leger1999automated}
Chris Leger and John Bares.
\newblock Automated task-based synthesis and optimization of field robots.
\newblock 1999.

\bibitem{whitman2020modular}
Julian Whitman, Raunaq Bhirangi, Matthew~J Travers, and Howie Choset.
\newblock Modular robot design synthesis with deep reinforcement learning.
\newblock In {\em AAAI}, pages 10418--10425, 2020.

\bibitem{icer2017evolutionary}
Esra Icer, Heba~A Hassan, Khaled El-Ayat, and Matthias Althoff.
\newblock Evolutionary cost-optimal composition synthesis of modular robots
  considering a given task.
\newblock In {\em 2017 IEEE/RSJ International Conference on Intelligent Robots
  and Systems (IROS)}, pages 3562--3568. IEEE, 2017.

\bibitem{whitman2018task}
Julian Whitman and Howie Choset.
\newblock Task-specific manipulator design and trajectory synthesis.
\newblock {\em IEEE Robotics and Automation Letters}, 4(2):301--308, 2018.

\bibitem{kim1993formulation}
J-O Kim and Pradeep~K Khosla.
\newblock A formulation for task based design of robot manipulators.
\newblock In {\em Proceedings of 1993 IEEE/RSJ International Conference on
  Intelligent Robots and Systems (IROS'93)}, volume~3, pages 2310--2317. IEEE,
  1993.

\bibitem{liu2020optimizing}
Stefan~B Liu and Matthias Althoff.
\newblock Optimizing performance in automation through modular robots.
\newblock In {\em 2020 IEEE International Conference on Robotics and Automation
  (ICRA)}, pages 4044--4050. IEEE, 2020.

\bibitem{al2013task}
Hatem Al-Dois, AK~Jha, and RB~Mishra.
\newblock Task-based design optimization of serial robot manipulators.
\newblock {\em Engineering Optimization}, 45(6):647--658, 2013.

\bibitem{singh2018modular}
Satwinder Singh, Ashish Singla, and Ekta Singla.
\newblock Modular manipulators for cluttered environments: A task-based
  configuration design approach.
\newblock {\em Journal of Mechanisms and Robotics}, 10(5), 2018.

\bibitem{althoff2019effortless}
M~Althoff, A~Giusti, SB~Liu, and A~Pereira.
\newblock Effortless creation of safe robots from modules through
  self-programming and self-verification.
\newblock {\em Science Robotics}, 4(31):eaaw1924, 2019.

\bibitem{corke2007simple}
Peter~I Corke.
\newblock A simple and systematic approach to assigning denavit--hartenberg
  parameters.
\newblock {\em IEEE transactions on robotics}, 23(3):590--594, 2007.

\bibitem{camposautomated2020}
Thais Campos, Samhita Marri, and Hadas Kress-Gazit.
\newblock Automated synthesis of modular manipulators’ structure and control
  for continuous tasks around obstacles.
\newblock In {\em Robotics Science and Systems}, 2020.

\bibitem{siciliano2016springer}
Bruno Siciliano and Oussama Khatib.
\newblock {\em Springer handbook of robotics}.
\newblock Springer, 2016.

\bibitem{siciliano2010robotics}
Bruno Siciliano, Lorenzo Sciavicco, Luigi Villani, and Giuseppe Oriolo.
\newblock {\em Robotics: modelling, planning and control}.
\newblock Springer Science \& Business Media, 2010.

\bibitem{corke2017robotics}
Peter Corke.
\newblock {\em Robotics, vision and control: fundamental algorithms in
  MATLAB{\textregistered} second, completely revised}, volume 118.
\newblock Springer, 2017.

\bibitem{Shortest2020}
Nick.
\newblock Shortest2020 distance between two line segments.
\newblock MATLAB Central File Exchange, Dec. 16 2020. [Online].

\bibitem{geom3d2020}
David Legland.
\newblock geom3d2020.
\newblock MATLAB Central File Exchange, Dec. 16 2020. [Online].

\bibitem{rrtmatlab2019}
Adnan Munawar.
\newblock Matlab implementation of rrt variants.
\newblock \url{https://github.com/adnanmunawar/matlab-rrt-variants}, 2019.

\bibitem{liu2000performance}
Xin-Jun Liu, Jinsong Wang, and Feng Gao.
\newblock Performance atlases of the workspace for planar 3-dof parallel
  manipulators.
\newblock {\em Robotica}, 18(5):563--568, 2000.

\end{thebibliography}

\end{document}